\documentclass[lettersize,journal]{IEEEtran}
\usepackage{amsmath,amsfonts}
\usepackage{algorithmic}
\usepackage{algorithm}
\usepackage{array}
\usepackage[caption=false,font=normalsize,labelfont=sf,textfont=sf]{subfig}
\usepackage{textcomp}
\usepackage{stfloats}
\usepackage{url}
\usepackage{verbatim}
\usepackage{graphicx}
\usepackage{cite}

\usepackage{booktabs}
\usepackage{multirow}
\usepackage{xcolor}
\usepackage{tikz}
\usepackage{pgfplots}
\pgfplotsset{compat=newest}

\begin{document}

\title{Moir\'eXNet: Adaptive Multi-Scale Demoir\'eing with Linear Attention Test-Time Training and Truncated Flow Matching Prior}

\author{Liangyan Li, Yimo Ning,  Kevin Le, Wei Dong, Yunzhe Li, Jun Chen,~\IEEEmembership{Senior Member,~IEEE,}, Xiaohong Liu,~\IEEEmembership{Senior Member,~IEEE,}
\thanks{This paper was produced by the IEEE Publication Technology Group. They are in Piscataway, NJ.}
\thanks{Manuscript received April 19, 2021; revised August 16, 2021.}}

\markboth{Journal of \LaTeX\ Class Files,~Vol.~14, No.~8, August~2021}%
{Shell \MakeLowercase{\textit{et al.}}: A Sample Article Using IEEEtran.cls for IEEE Journals}

\IEEEpubid{0000--0000/00\$00.00~\copyright~2021 IEEE}

\maketitle

\begin{abstract}

This paper introduces a novel framework for image and video demoiréing by integrating Maximum A Posteriori (MAP) estimation with advanced deep learning techniques. Demoiréing addresses inherently nonlinear degradation processes, which pose significant challenges for existing methods. 
Traditional supervised learning approaches either fail to remove moiré patterns completely or produce overly smooth results. This stems from constrained model capacity and scarce training data, which inadequately represent clean image distribution and hinder accurate reconstruction of ground-truth images. While generative models excel in image restoration for linear degradations, they struggle with nonlinear cases such as demoiréing and often introduce artifacts.  

To address these limitations, we propose a hybrid MAP-based framework that integrates two complementary components. The first is a supervised learning model enhanced with efficient linear attention Test-Time Training (TTT) modules, which directly learn nonlinear mappings for RAW-to-sRGB demoiréing. The second is a Truncated Flow Matching Prior (TFMP) that further refines the outputs by aligning them with the clean image distribution, effectively restoring high-frequency details and suppressing artifacts. These two components combine the computational efficiency of linear attention with the refinement abilities of generative models, resulting in improved restoration performance.


\end{abstract}


\section{Introduction}
\label{sec:intro}
\IEEEPARstart{M}{oir\'e} patterns, caused by interference between grid-like structures such as camera sensors and LED screens~\cite{wang2023coarse}, are visually disruptive artifacts characterized by wavy lines, ripples, or colorful distortions~\cite{sun2018moire}. These patterns degrade image quality and are challenging to remove due to their complex, content-dependent variations in thickness, frequency, and color, which often blend with fine image details~\cite{zheng2021learning}.  

\begin{figure}[t]
    \centering
    \includegraphics[width=\columnwidth]{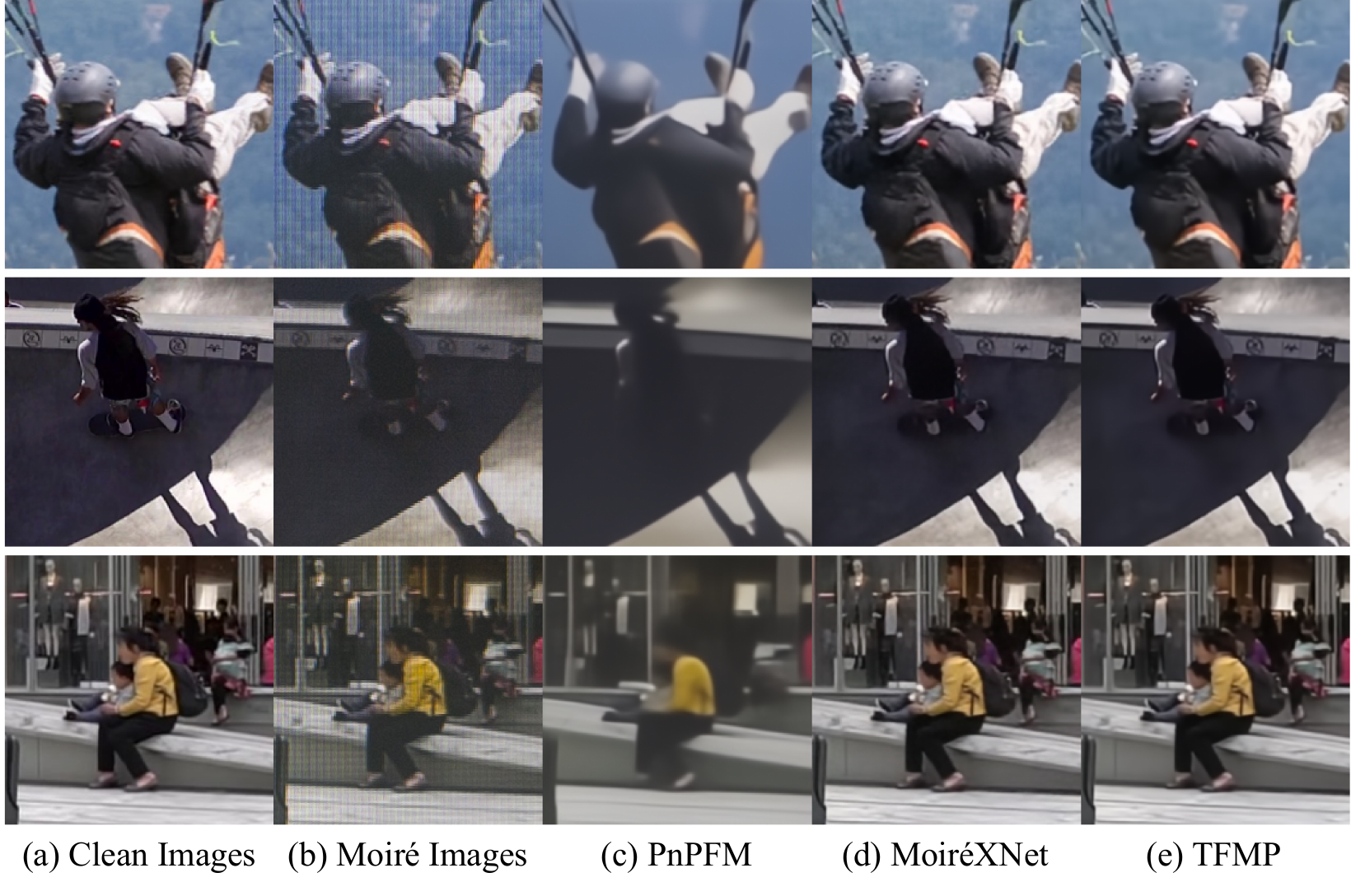}
    \caption{Visual comparison of moir\'e artifact removal and detail preservation: (a) Clean Images, (b) Moir\'e Images, (c) PnP Flow Matching with Moir\'e sRGB as inputs, (d) Moir\'eXNet results (ours), and (e) Moir\'eXNet results enhanced refinement via TFMP.
    Using pretrained PnP Flow Matching with a linear kernel on moir\'e sRGB inputs (d) leads to artifacts like bullring effects due to the nonlinear nature of the moir\'e pattern. Our framework (d) achieves superior structural fidelity and artifact suppression while preserving high-frequency details. Refining the results of Moir\'eXNet with TFMP (e) achieves further performance enhancements. 
   }
    \label{fig:pnp_result}
    \vspace{-0.4cm}
\end{figure}
Conventional demoir\'eing methods, relying on classical filters or signal decomposition models~\cite{siddiqui2009hardware,sun2014scanned,yang2017textured,kim2008integral,wei2012median}, struggle to handle the non-linear and intricate nature of moir\'e artifacts. The introduction of paired real~\cite{sun2018moire,he2020fhde,yue2022recaptured,yu2022towards,zheng2020image,peng2024image,dai2022video,yue2023recaptured} and synthetic datasets~\cite{zhong2024learning} has enabled supervised learning methods~\cite{sun2018moire,liu2018demoir,yuan2019aim2019challengeimage,cheng2019multi,he2019mop,gao2019moire,he2020fhde,liu2020wavelet,zheng2020image,Liu_2020_CVPR_Workshops,zheng2021learning,wang2021image,yu2022towards,niu2023progressive,wang2023coarse} to achieve notable success in recovering clean sRGB images from corrupted inputs. However, these methods are limited by their reliance on finite training datasets, which fail to capture the true distribution of clean images. This limitation, compounded by the non-linear transformations in the Image Signal Processor (ISP) pipeline, often results in oversmoothed outputs with missing high-frequency details. While some efforts have been made to incorporate frequency-domain information~\cite{he2020fhde,liu2020wavelet,dai2024freqformer,zheng2020image,wang2023coarse} or utilize RAW domain data~\cite{yue2022recaptured,yue2023recaptured,xu2024image}, they still fall short of accurately recovering fine textures and edges, leading to suboptimal restoration quality.
\begin{figure*}[!ht]
\centering
\includegraphics[width=1\linewidth, page=1]{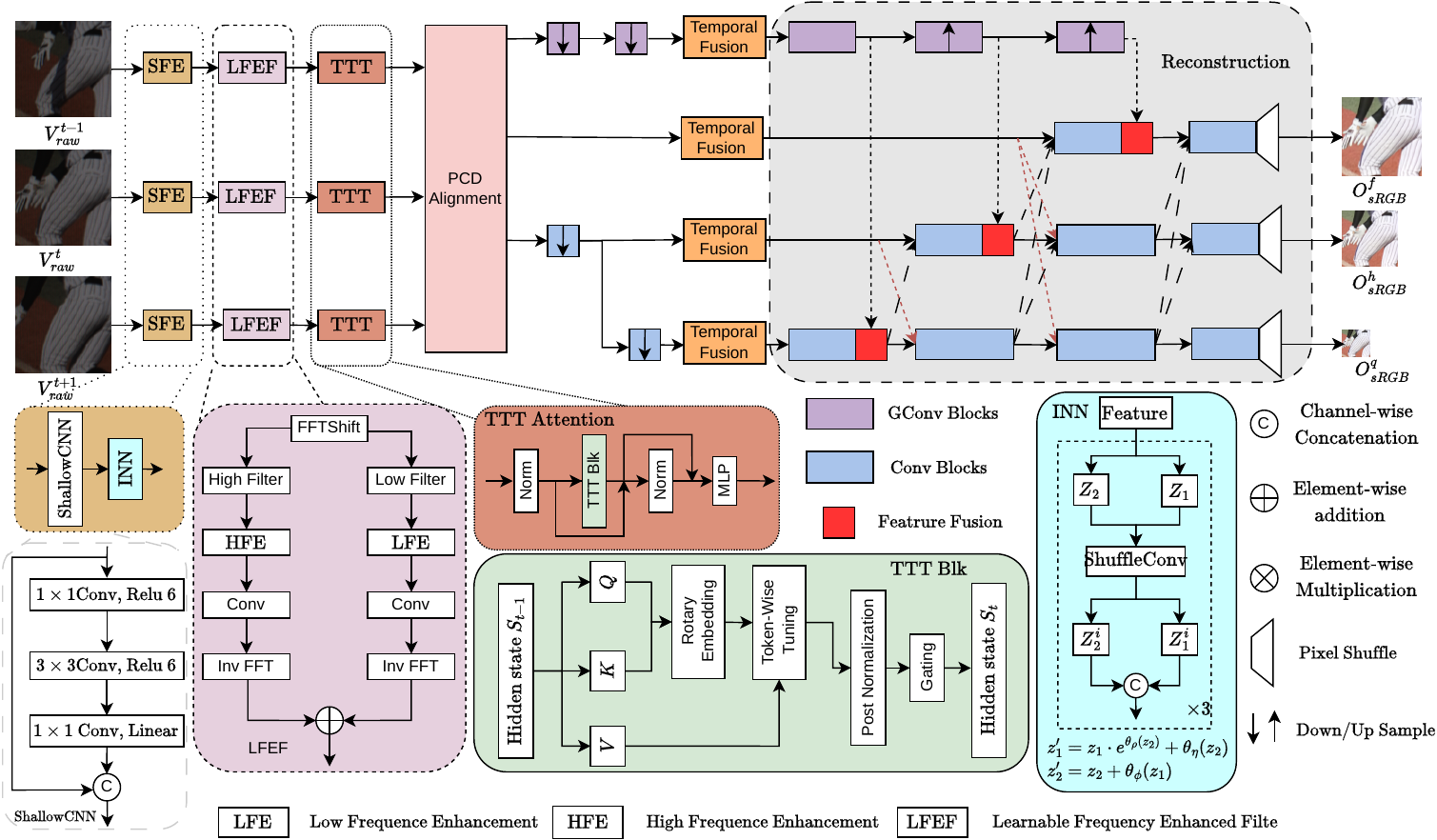}
\caption{An overview of the proposed method.}
\label{fig:architecture}
\end{figure*}

Generative models~\cite{goodfellow2014generativeadversarialnetworks,kingma2022autoencodingvariationalbayes,ho2020denoisingdiffusionprobabilisticmodels,marinescu2020bayesian} have shown strong performance in image restoration tasks by leveraging learned priors to recover missing details. For tasks such as denoising, deblurring, and super resolution, plug-and-play (PnP) denoisers~\cite{bora2017compressed,asim2020invertible,wei2022deep,altekruger2023patchnr,ben2024d,zhang2025flow,pokle2023training,chung2022diffusion,song2023pseudoinverse,hurault2022proximal,liu2021recovery,martin2024pnp} are widely adopted for reconstruction. However, these approaches predominantly assume \textit{linear} degradation processes (e.g., additive noise, known blur kernels, uniform downsampling). Moir\'e patterns, in contrast, pose a fundamentally different and more complex challenge. They arise from nonlinear interactions between scene textures and sensor sampling grids, resulting in spatially varying aliasing effects that resist closed-form characterization. This inherent nonlinearity limits generative models' ability to disentangle artifacts from true image content, often leading to residual artifacts or hallucinated details in restored images, as illustrated in Figure~\ref{fig:pnp_result}, column (c). MRGAN~\cite{yue2021unsupervised}, an unsupervised approach based on CycleGAN~\cite{Zhu_2017_ICCV}, demonstrates progress in moir\'e removal by training generators with self-supervised techniques. However, it fails to fully exploit the benefits of supervised learning and clean image priors, which restricts its effectiveness.

This paper proposes a generic approach to tackling image/video restoration and demonstrates its effectiveness, particularly in the challenging task of demoir\'eing. Our contributions can be summarized as follows:

\begin{enumerate}

\item \textbf{Hybrid MAP-based framework}: We introduce a novel framework that combines supervised learning with generative priors to address the nonlinear and non-stationary nature of moir\'e degradation.
\item \textbf{Efficient Test-Time Training (TTT) modules}: We incorporate linear attention TTT modules into a supervised model, enabling efficient and robust RAW-to-sRGB demoir\'eing through direct nonlinear mappings.
\item \textbf{Flow Matching generative prior}: We leverage TFPM to refine restoration outputs, effectively aligning them with the clean image distribution to recover high-frequency details and suppress artifacts.
\item \textbf{State-of-the-art performance}: Our approach demonstrates superior results on benchmark datasets, achieving significant improvements in quantitative metrics (e.g., PSNR) and visual quality over prior methods.
\end{enumerate}

\section{Related Works}
\label{sec:related}
\subsection{Moir\'e Pattern Removal}

Moir\'e patterns result from the interference of similar frequencies, degrading the quality of screen captures. A moir\'e pattern remover restores clean images or videos by eliminating these patterns and correcting color deviations. Conventional methods~\cite{sasada2003stationary,siddiqui2009hardware} primarily focus on specific types of moir\'e patterns. In contrast, supervised learning-based image demoir\'eing approaches excel at learning diverse moir\'e patterns. This progress has been driven by the availability of high-quality moir\'e datasets~\cite{yue2022recaptured,zheng2020image,yu2022towards,sun2018moire,hefhde2net,dai2022video,yue2023recaptured} and advancements in deep learning backbones~\cite{he2016deep,vaswani2017attention,liu2025vmamba,sun2024learning}.   

For image demoir\'eing, most approaches utilize Convolutional Neural Network (CNN)-based architectures, integrating multiscale features~\cite{liu2018demoir,sun2018moire,cheng2019multi,he2019mop,zheng2020image,9022550,9151073,vien2020dual}, attention mechanisms~\cite{kim2020c3net,xu2020moire} (e.g., channel, spatial, and color) and frequency-domain techniques~\cite{9022550,zheng2020image,he2020fhde,liu2020wavelet,vien2020dual,luo2020deep,zheng2021learning}   to tackle the complex patterns of moir\'e artifacts. 3DNet~\cite{wang2021image} leverages both spatial- and frequency-domain knowledge through a dual-domain distillation network. DDA~\cite{zhang2023real} focuses on efficient image demoir\'eing for real-time applications.
The aforementioned methods predominantly operate in the sRGB domain, where the ISP discards much of the original sensor information through processes such as tone mapping, white balance, and compression. In contrast, RAW-domain images retain unprocessed sensor data, preserving richer details and naturally exhibiting reduced moir\'e patterns.
RDNet~\cite{yue2022recaptured} introduces the first RAW-domain demoir\'eing dataset, leveraging the richer information available in RAW images and incorporating a multi-scale encoder with multi-level feature fusion. However, despite employing a class-specific learning strategy to handle different types of screen content, it lacks flexibility when applied to diverse scenes, ultimately limiting its generalization capability.  Studies such as~\cite{yue2022recaptured, xu2025image, xu2024image} explore image demoir\'eing in both RAW and sRGB domains. However, these methods struggle to handle diverse and complex scenarios effectively.


For video demoiréing, VDmoir\'e~\cite{dai2022video} introduces the first dedicated dataset and  a baseline model, while RawVDemoir\'e~\cite{yue2023recaptured} proposes a temporal alignment method specifically for RAW video demoir\'eing.
Compared to their image demoir\'eing counterparts, which primarily focus on feature extraction and fusion, video demoir\'eing  methods~\cite{xu2024direction, niu2024std, liu2024video, cheng2023recaptured, quan2023deep, dai2022video} emphasize leveraging temporal information from neighboring frames and aggregating multi-frame features to enhance the quality of the restored video frames.

A key challenge in both image and video reconstruction lies in extracting rich features that can effectively capture spatial and temporal dependencies.
Transformers~\cite{vaswani2017attention,dosovitskiy2020image} have consistently outperformed CNN-based models~\cite{simonyan2014very,he2015deep} across a wide variety of tasks~\cite{wang2023transfer,lyu2022multimodal,lyu2024backdooring,Chen_2023_CVPR,9819903,Liu_2019_ICCV,lyu2023attention,Liu_2023_CVPR,wang2022work,wang2023transfer,wang2024exploring}, thanks to their ability to capture global dependencies through self-attention mechanisms effectively. Such mechanisms rely on the Key-Value (KV) cache to store historical context, with the attention output at time $t$  given by:
\[
z_t = \text{softmax}\left(\frac{Q K^\top}{\sqrt{d_k}}\right) V,
\]
where the $\text{softmax}$ operation computes the attention weights. Self-attention explicitly stores all historical context, resulting in memory requirements that grow linearly with the sequence length $O(t)$  and computational complexity of  $O(t^2)$ due to pairwise interactions between all tokens. In contrast, Mamba RNNs~\cite{xu2024demmamba,gu2023mamba} reduce complexity by approximating the key and value matrices using compression matrices $W_k$ and $W_v$, yielding $K' = W_k K$, $V' = W_v V$. The attention output then becomes:
\[
z_t = \text{softmax}\left(\frac{Q {K'}^\top}{\sqrt{d_k}}\right) V',
\]
where $K'$ and $V'$ are compressed representations of  $K$ and $V$. This approach reduces computational cost to $O(t)$, improving efficiency at the cost of some expressiveness.

More recently, TTT blocks~\cite{sun2024learning,zhao2023cddfuse} have emerged as a method to bridge the gap between Transformers and RNNs by employing efficient parametric updates. TTT avoids maintaining a growing KV cache by using a parametric hidden state $s_t$, which is updated as follows: \[
s_t = f(s_{t-1}, x_t; W),
\]
where $s_{t-1}$ is the previous hidden state, $x_t$ is the current token input, and $W$ represents the learned parameters. The output is generated as:
\[
z_t = g(s_t; W).
\]
This hidden state is iteratively updated at each time step, representing a compressed summary of all previous tokens. TTT does not maintain an explicit KV cache, resulting in a fixed-size representation with $O(1)$ memory requirements. While TTT is highly efficient for processing long sequences, it is typically less expressive than full self-attention mechanisms. In this work, we adopt TTT linear attention layers as our primary building blocks, leveraging their long-range attention capabilities while maintaining computational efficiency.

However, studies~\cite{tang2023learning, wang2022anti} have shown that self-attention behaves like a low-pass filter—favoring low-frequency (coarse) components while suppressing high-frequency (fine-grained) details. To address this limitation and better preserve high-frequency information, we introduce two architectural enhancements before the TTT blocks: Invertible Neural Networks (INNs)~\cite{dinh2016density, kingma2018glow, dinh2014nice} for efficient, lossless feature transformation, and a Learnable Frequency Enhanced Filter (LFEF) to adaptively amplify both low- and high-frequency components.


Detailed architecture is presented in Section~\ref{subsec:demoireing}, and performance evaluations are reported in Section~\ref{sec:exp}. As shown in Table~\ref{table:RAWVDM}, the TTT linear attention layers offer low computational complexity and fast inference, while maintaining high restoration quality, as demonstrated by competitive PSNR and SSIM scores.

\subsection{Pretrained Image Priors for Image Restoration}

Although traditional supervised image demoir\'eing methods achieve high PSNR~\cite{wang2004image}, their results often suffer from oversmoothing artifacts~\cite{isola2017image}. This limitation stems from two key factors: (1) Euclidean distance-based loss functions in neural networks prioritize pixel-wise fidelity at the expense of perceptual quality, and (2) constrained dataset diversity restricts the model's capacity to learn accurate image mappings. 

To address these challenges, recent approaches have integrated prior knowledge of natural image statistics into restoration pipelines. Image restoration is typically formulated as an inverse problem, where the goal is to recover a clean image $x$ from noisy observations $y$ based on the forward model: $y = Hx + n$, where $H$ is the forward operator, which is typically a \textit{linear} operation (e.g., a blurring matrix or downsampling operator), and $n$ represents additive noise, often assumed to follow a Gaussian distribution. One common approach to solving this inverse problem is to optimize a regularized objective function:
\begin{equation}
    \hat{x} = \arg\min_x \frac{1}{2} \| Hx - y \|_2^2 + \lambda \Phi(x).
\end{equation}

The first term, $\frac{1}{2} \| Hx - y \|_2^2$, enforces consistency with the observed data (data fidelity term), while
the second term, $\lambda \Phi(x)$, encodes prior knowledge about the image (regularization term), such as smoothness or sparsity. Traditional hand-craft methods~\cite{roth2005fields,zhu1997prior,geman1984stochastic,he2010single,rudin1992nonlinear} relied on explicit mathematical models of natural image statistics, such as Fourier spectrum~\cite{ruderman1993statistics}, total variation (TV)~\cite{rudin1992nonlinear,chambolle1998nonlinear,donoho2002noising,donoho1995adapting,moulin1999analysis}, sparsity priors~\cite{mallat1999wavelet,chan2005recent,chambolle2004algorithm,rudin1992nonlinear}, and patch-based Gaussian mixtures priors~\cite{zoran2011learning,liu2021recovery} to guide the restoration process.
These pre-designed priors have chanllenge in terms of real image restoration tasks, such as image demoi\'ering. 
Handcrafted priors rely on fixed assumptions about the source of the moiré. When applied to new types of moiré (e.g., fabric textures vs. screen displays), they often fail.

The Plug-and-Play (PnP)~\cite{venkatakrishnan2013plug,sreehari2016plug} framework revolutionized image restoration by decoupling the prior from the forward model. Instead of explicitly defining a regularization term $\Phi(x)$, PnP leverages powerful image denoising algorithms as implicit priors. For example, advanced nonlearned denoisers like BM3D~\cite{chan2016plug} and CNN-based denoiser~\cite{zhang2017beyond,arjomand2017deep,chen2016trainable,lunz2018adversarial,meinhardt2017learning,zhang2021plug}
have been used for this purpose. Many modern learned priors exploit the capabilities of generative models, which excel at capturing complex natural image distributions.  Generative models such as GANs~\cite{goodfellow2014generativeadversarialnetworks}, VAEs~\cite{kingma2022autoencodingvariationalbayes}, diffusion models~\cite{song2020score,ho2020denoisingdiffusionprobabilisticmodels,sohl2015deep}, and normalizing flows~\cite{dinh2016density, kingma2018glow} have shown immense potential in this regard, making them valuable tools for modeling priors~\cite{bora2017compressed,asim2020invertible,wei2022deep,altekruger2023patchnr,ben2024d,zhang2025flow,pokle2023training,chung2022diffusion,song2023pseudoinverse}.
Diff-Plugin~\cite{liu2024diff} enhances pretrained DDPMs for image restoration by adding a lightweight plugin to refine details and improve task-specific outputs.
A recent extension of normalizing flows, known as flow matching~\cite{lipman2022flow,liu2022flow}, optimizes transport paths between distributions and has been explored for various image restoration tasks within PnP frameworks~\cite{martin2024pnp}.

Even though PnP denoisers excel in addressing linear degradation tasks such as denoising, deblurring, and super-resolution, effectively balancing fidelity and perceptual quality, their effectiveness is significantly  constrained in handling nonlinear degradation tasks like demoir\'eing or JPEG artifact removal. As shown in Fig.~\ref{fig:pnp_result}, when applied to image demoir\'eing, PnP-Flow~\cite{martin2024pnp} tends to introduce artifacts due to the inherent complexity of the degradation process. To overcome this limitation, we propose using a Truncated Flow Matching Prior (TFMP) as a refinement step to enhance the demoir\'eed images produced by a supervised model.

\section{Methodology}
\label{sec:method}
Moir\'e removal involves recovering a clean image \( x \) from a degraded observation \( y = M(x) + n \), where \( M(\cdot) \) represents a \textit{nonlinear}, scene-dependent degradation caused by interference between high-frequency textures and sampling grids. In this section, we present our hybrid approach: A supervised learning model in Section~\ref{subsec:demoireing}, enhanced with efficient linear attention TTT modules, directly learns nonlinear mappings for RAW-to-sRGB demoir\'eing. This stage incorporates INN and LFEF modules to refine features in both the spatial and frequency domains, effectively removing coarse patterns while preserving structural content. 
 A truncated flow matching model in Section~\ref{subsec:pnp_FMF}  further refines the outputs by aligning them with the clean image distribution. This step restores high-frequency details and suppresses residual artifacts through distribution matching, enabling photorealistic texture recovery.

\subsection{Demoir\'eing Network Architecture}
\label{subsec:demoireing}

Our demoir\'eing framework builds upon the VDRaw framework~\cite{cheng2023recaptured}, replacing the convolutional layers in the preprocessing phase with a combination of ShallowCNN and an INN module to enhance high-frequency detail preservation. Furthermore, we adaptively combine the frequency domain features by learning a weighted filter before applying TTT linear attention for multi-scale feature extraction. In our model, we selected the TTT 1B version and adjusted the hidden size to 256.

Fig.~\ref{fig:architecture} provides an overview of our proposed \textbf{Moir\'eXNet} model for video demoir\'eing.
More specifically, the key stages include  Shallow Feature Extraction (SFE), Deep Feature Extraction (DFE), Auxiliary Frames Alignment and Blend (AFAB) and Hierarchical Reconstruction (HR). The model takes three neighboring RAW images, $\mathbf{V}^i_{raw} \in \mathbb{R}^{\frac{H}{2} \times \frac{W}{2} \times 4}$, where $i \in \{t-1, t, t+1\}$, in the 4-channel RGGB format as input and directly reconstructs the corresponding central frame sRGB image $\mathbf{O}^t_{sRGB}$. For clarity and consistency, we adopt the same notation as  VDRaw. 

The SFE block is designed to extract shallow features, denoted as $\mathbf{F}_{\text{raw}}$, from the RAW input data $\mathbf{V}_{\text{raw}}$. The SFE block uses lightweight convolutional layers to embed raw input frames (reference + auxiliary frames) into an initial feature space. Then, an INN module, as proposed in~\cite{zhao2023cddfuse}, further ensures lossless information preservation by enabling the mutual reconstruction of input and output features. This unique property allows the INN to serve as a lossless feature extraction mechanism, ensuring that all critical information is retained for the subsequent DFE process.
The DFE block generates multi-scale features by downsampling shallow features using bilinear interpolation with scaling factors of 0.5 and 0.25. At each level, the combined LFEF and TTT Linear Attention blocks extract deep features from each frame. The LFEF addresses the limitations of attention layers, which tend to lose high-frequency information, by extracting richer features and enhancing overall model performance. Detailed architecture is shown in Figure~\ref{fig:architecture}.

At each scale, TTT linear attention blocks capture long-range dependencies to enhance global context understanding while maintaining fine-grained details. The multi-scale features are finally fused using the VDRaw strategy, ensuring structural consistency and effective feature integration.

We have enhanced the feature extraction blocks compared to the VDRaw framework, while keeping the feature alignment and reconstruction blocks unchanged. Following the approach in~\cite{dai2022video, wang2019edvr, xu2024image}, we employ pyramid cascading deformable (PCD) alignment to align the features of $\mathbf{V}^{t-1}_{raw}$
and $\mathbf{V}^{t+1}_{raw}$ with those of $\mathbf{V}^{t}_{raw}$. For reconstruction, we adhere to the original VDRaw framework, where the final output consists of three different resolutions of sRGB images: full resolution, half resolution, and quarter resolution $\mathbf{O}^f_{sRGB}$, $\mathbf{O}^h_{sRGB}$, 
$\mathbf{O}^q_{sRGB}$, which are used to calculate the multiscale loss.

For image demoir\'eing tasks, features are extracted from a single input frame, bypassing the need for the PCD module. Instead, multi-scale features are fused using TTT linear attention blocks before being passed to the reconstruction backbone for image restoration.

\subsection{Flow Matching for Iterative Refinement of Degraded Images}
\label{subsec:pnp_FMF}
We propose an iterative refinement process leveraging a TFMP to enhance the outputs of the Moir\'eXNet model. The denoiser learns a velocity field that maps degraded images to clean ones, enabling a stepwise recovery that progressively brings images closer to the ground truth. We initialize the iterative refinement process with a degraded image \(\Tilde{x}\), which serves as the initial estimate of the clean image \(x\), as \(\Tilde{x}\) is significantly closer to \(x\) compared to the moir\'eimage \(y\). Thus, we set \(x_t = \Tilde{x}\), with \(t\) starting from a higher value (e.g., \(t = 0.95\)) rather than 0. Here, \(t \in [0.95, 1]\)  represents the progression through the refinement process, with five samples drawn at each step. A refined version of the Moir\'eXNet model's output is obtained by applying a few iterations of the denoiser.

The flow matching model learns a velocity field $\frac{\partial x_t}{\partial t} = v(x_t, t)$, which defines the gradient direction guiding the degraded image toward the clean image at timestep \( t \). In our approach, this velocity field is iteratively applied to refine the Moir\'eXNet model's output by updating \( x_t \) as follows:
\[
x_{t-1} = x_t + \Delta t \cdot v(x_t, t),
\]
where \( \Delta t \) is the step size.

The flow matching model is pretrained to learn the transformation dynamics from degraded images to clean images, ensuring robust refinement. The iterative process progressively improves the image, making it cleaner and closer to the ground truth. The method works for a wide variety of image degradation types, such as noise, blur, and compression artifacts.

\section{Experiments}
\label{sec:exp}
\begin{table*}[htbp]
       \footnotesize
	\centering
	\caption{A comparison of state-of-the-art methods for image and video demoir\'eing on RawVDemoir\'e dataset, evaluated using average PSNR, SSIM, LPIPS, and computational complexity. The best results are bolded in red, while the second-best results are bolded in black. This table highlights the state-of-the-art performance of our model, Moir\'eXNet, in both image and video demoir\'eing tasks.}
    \begin{tabular}{ccccccc}
		\toprule
            & \textbf{Method} & 
            \textbf{Input type} &
            \textbf{PSNR}$\uparrow$ & \textbf{SSIM}$\uparrow$ & \textbf{LPIPS}$\downarrow$ & \textbf{Inference time (s)} \\
            \midrule
		\multirow{2}*{\textbf{Image}} & RDNet~\cite{yue2022recaptured}  & RAW& 25.892 & 0.8939 & 0.1508 & 2.514 \\
            & RRID~\cite{xu2025image} & sRGB+RAW& \textbf{27.283} & \textbf{0.9029} & \textbf{0.1168} & \textcolor{black}{\textbf{0.501}} \\
            & \textbf{Moir\'eXNet} & RAW& \textcolor{red}{\textbf{29.590}} & \textcolor{red}{\textbf{0.9170}} & \textcolor{red}{\textbf{0.0936}} & \textcolor{red}{\textbf{0.070}} \\
            \midrule
            \multirow{9}*{\textbf{Video}} 
            & VDMoir\'e~\cite{dai2022video} &sRGB+RAW& 27.277 & 0.9071 & 0.1044 & 1.057 \\
            & VDMoir\'e$\ast$ &sRGB+RAW & 27.747 & 0.9116 & 0.0995 & 1.125\\
            & DTNet~\cite{xu2024direction} &sRGB & 27.363 & 0.8963 & 0.1425 & 0.972 \\
            & DTNet$\ast$ &sRGB & 27.892 & 0.9055 & 0.1135 & 1.050 \\
            & VDRaw~\cite{cheng2023recaptured} &sRGB+RAW & 28.706 & 0.9201& 0.0904 & 1.247 \\
            & DemMamba~\cite{xu2024demmamba} &RAW& 30.004 & 0.9169 & \textbf{0.0901} & \textbf{\textcolor{black}{0.446}} \\
            & \textbf{Moir\'eXNet} & RAW & \textbf{30.127} & \textbf{0.9258} & \textcolor{red}{\textbf{0.0847}} & \textcolor{red}{\textbf{0.070}} \\
            & \textbf{TFMP} & RAW & \textcolor{red}{\textbf{30.214}} & \textcolor{red}{\textbf{0.9281}} & 0.0973 & \textcolor{black}{-} \\
		\bottomrule
\end{tabular}
\label{table:RAWVDM}
\end{table*}

We compare the proposed Moir\'eXNet and its refined version, TFMP, with state-of-the-art methods and evaluate their performance on both video and image demoir\'eing tasks.
\subsection{Experimental Setup}

\begin{table*}[htbp]
    \centering
    \caption{Quantitative comparison with the state-of-the-art demoir\'eing  approaches and RAW image restoration methods on \textbf{TMM22 dataset} \cite{yue2022recaptured} in terms of average PSNR, SSIM, LPIPS and computational complexity. The best results are bolded in red, while the second-best results are bolded in black.}
    \label{table:quantitative-rawid}
    \scalebox{0.83}{
        \begin{tabular}{ccccccccccc}
        \toprule
        \multirow{3}*{Index} & \multicolumn{10}{c}{Methods}\\
		\cmidrule(r){2-11}
		 ~ & DMCNN & WDNet  & ESDNet & RDNet 
  & VDRaw& CR3Net & RRID & DemMamba & Moir\'eXNet & TFMP\\
         ~ & \cite{sun2018moire} & \cite{liu2020wavelet} & \cite{yu2022towards} & \cite{yue2022recaptured} 
 & \cite{cheng2023recaptured} & \cite{song2023real} &~\cite{xu2025image}&~\cite{xu2024demmamba}&Ours & Ours\\
        \midrule
        \# Input type & sRGB & sRGB & sRGB & RAW 
 & RAW & sRGB+RAW & sRGB+RAW &RAW  & RAW & RAW\\
        \midrule
        PSNR$\uparrow$ & 23.54 & 22.33 & 26.77 & 26.16 
 & 27.26 & 23.75 &  27.88 & \textbf{\textcolor{red}{28.14}} & 28.03 & \textbf{28.08}\\
        SSIM$\uparrow$ & 0.885 & 0.802 &  0.927 & 0.921 
 & 0.935 & 0.934 & \textbf{\textcolor{red}{0.938}} &0.936 & \textbf{0.937} & \textbf{\textcolor{red}{0.938}}\\
        LPIPS$\downarrow$ & 0.154 & 0.166 & 0.089 & 0.091
 & 0.075 & 0.102 & 0.079 & \textbf{0.067} & \textbf{\textcolor{red}{0.066}} & \textbf{0.067}\\
        \cmidrule(r){1-11}
        \end{tabular}}
\end{table*}

\noindent\textbf{Training Details.}
Our experiments are conducted on a machine equipped with two NVIDIA A100 GPUs. 
We train our methods using the AdamW optimizer with an initial learning rate of $3 \times 10^{-4}$, betas (0.9, 0.999) for momentum and variance smoothing, and weight decay to improve generalization. The learning rate is adjusted using the ReduceLROnPlateau scheduler, which reduces it by a factor of 0.8 if validation loss does not improve for 3 consecutive epochs, with a minimum learning rate of $5 \times 10^{-6}$. This setup ensures efficient and stable training with adaptive learning rate adjustments. We begin  training with L1 VGG loss for 175 epochs, followed by fine-tuning with wavelet loss for an additional 41 epochs to further refine the results.

\subsection{Datasets}
\begin{figure*}[!htbp]
\captionsetup{skip=2pt}
\centering
\includegraphics[width=1\linewidth]
{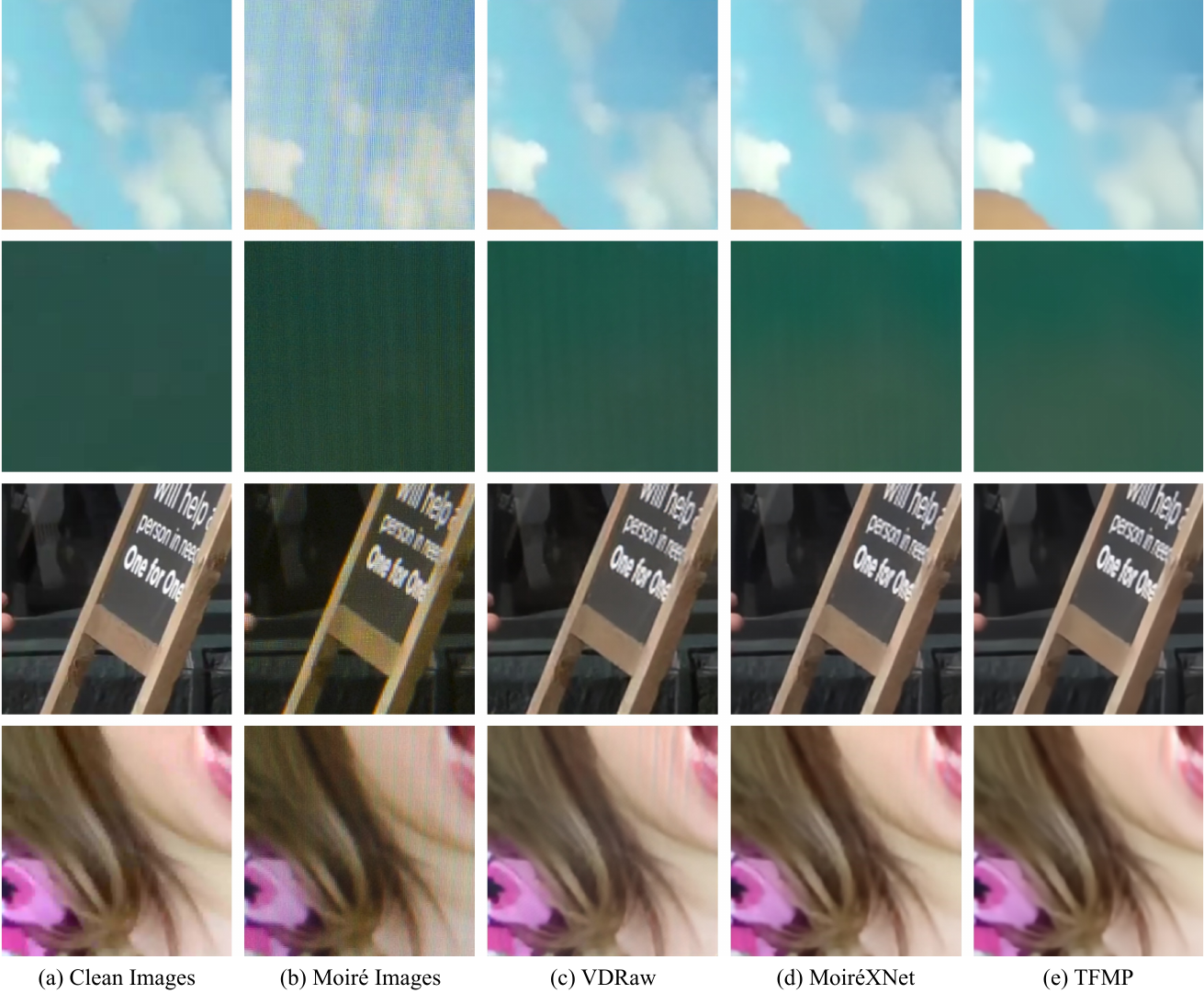}
\caption{Qualitative comparison on RAW video demoir\'eing  RawVDemoir\'e~\cite{cheng2023recaptured}.}
\label{fig:vdmoire}
\end{figure*}

\begin{figure*}[!ht]
\captionsetup{skip=2pt}
\centering
\includegraphics[width=1\linewidth]{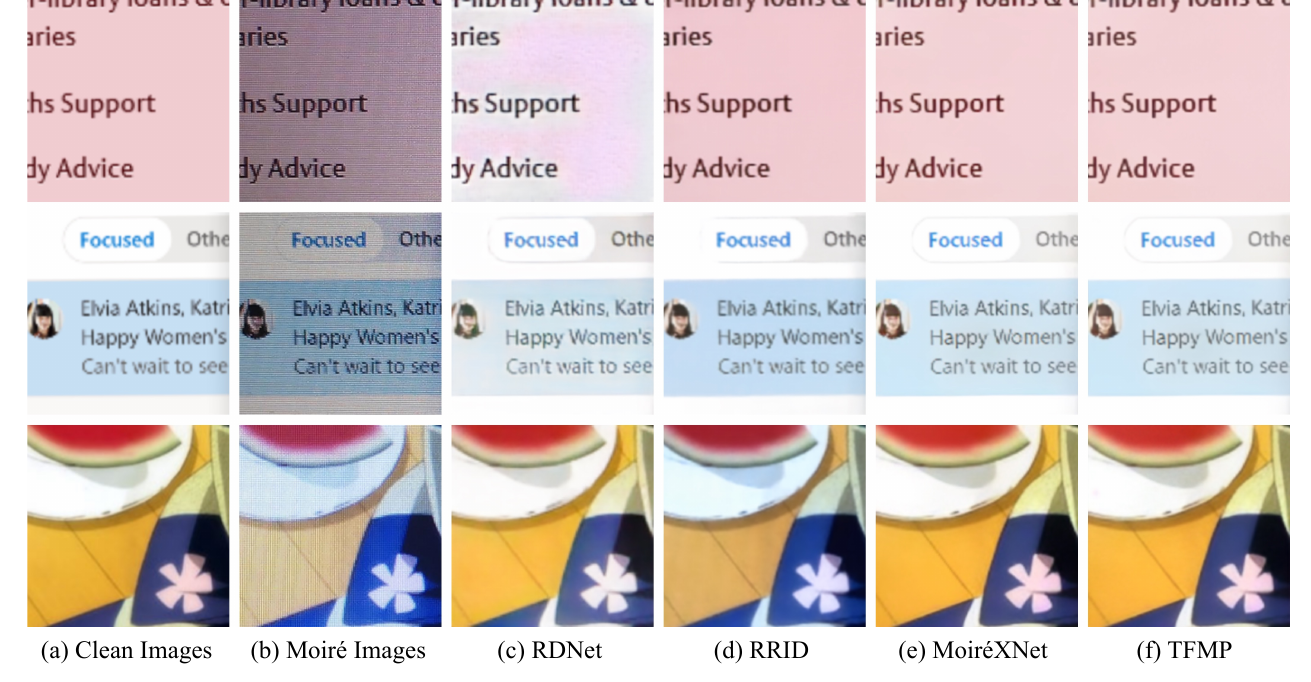}
\caption{Qualitative comparison on RAW image demoir\'eing  TMM22 dataset~\cite{yue2022recaptured}.}
\label{fig:TMM22}
\end{figure*}

For the RAW-domain image and video demoir\'eing task, we conduct experiments using the RawVDemoir\'e dataset~\cite{cheng2023recaptured} and the TMM22 dataset~\cite{yue2022recaptured}. The VDMoir\'e dataset includes 300 training videos and 50 testing videos, each with 60 frames at a resolution of 1080×720 (720p). For image demoir\'eing, the TMM22 dataset provides 540 RAW and sRGB image pairs for training and 408 pairs for testing, with image patches cropped to 256×256 for training and 512×512 for testing. In both datasets, RAW moir\'e inputs are compared against sRGB ground truth images to evaluate the performance of the proposed method.  

\subsection{Loss Function}
Relying solely on pixel-wise losses in the sRGB domain, such as $L_1$ or $L_2$, is often insufficient. We combine $L_1$ loss and VGG-based perceptual loss as follows:
\begin{equation}
    \mathcal{L}_{\text{total}} = \lambda_{\text{vgg}} \mathcal{L}_{\text{vgg}}(I_{\text{pred}}, I_{\text{gt}}) + \lambda_{\ell_1} \mathcal{L}_{\ell_1}(I_{\text{pred}}, I_{\text{gt}}),
\end{equation}
where $\lambda_{\text{vgg}} = 0.3$ and $\lambda_{\ell_1} = 0.7$. 
The $L_1$ loss is:
\begin{equation}
    \mathcal{L}_{\ell_1} = \| I_{\text{pred}} - I_{\text{gt}} \|_1.
\end{equation}
The perceptual loss is computed using a pre-trained VGG-16~\cite{simonyan2014very} network:
\begin{equation}
    \mathcal{L}_{\text{vgg}} = \sum_{l \in \mathcal{F}} \| \phi_l (I_{\text{pred}}) - \phi_l (I_{\text{gt}}) \|_1,
\end{equation}
where $\phi_l$ represents the activation from layer $l$, and $\mathcal{F}$ is the set of feature layers.

\subsection{Evaluation}
For quantitative comparison, we use PSNR~\cite{wang2004image}, SSIM\cite{wang2004image}, and LPIPS~\cite{zhang2018unreasonable} to evaluate image quality. PSNR assesses pixel-level fidelity but overlooks structural and perceptual aspects. SSIM incorporates luminance, contrast, and structure, offering better alignment with human perception while remaining pixel-based. LPIPS leverages deep features to assess semantic distortions, enabling robust perceptual evaluation and broad use in demoir\'eing tasks~\cite{he2020fhde,zhang2018unreasonable,yu2022towards,xiao2024p}. Additionally, we evaluate model efficiency by reporting inference time for a comprehensive analysis.

Table~\ref{table:RAWVDM} compares the performance of various methods for image and video demoir\'eing based on PSNR, SSIM, LPIPS and inference time on RawVDemoir\'e dataset~\cite{yue2023recaptured}. The results demonstrate that Moir\'eXNet outperforms all other methods, excelling in both reconstruction quality and computational efficiency, and establishes a new benchmark for RAW
image and video demoir\'eing tasks. For image demoir\'eing, Moir\'eXNet achieves a PSNR of 29.590 dB, SSIM of 0.9170, and LPIPS of 0.0936, significantly outperforming RDNet~\cite{yue2022recaptured} and RRID~\cite{xu2025image}. Specifically, Moir\'eXNet's PSNR is +3.698 dB higher than RDNet (25.892 dB) and +2.307 dB higher than RRID (27.283 dB), while its SSIM is +0.0231 dB higher than RDNet (0.8939) and +0.0141 dB higher than RRID (0.9029). the LPIPS is  0.0572 lower than RDNet and 0.0232 lower than RRID. For video demoir\'eing, Moir\'eXNet achieves a PSNR of 30.127 dB, an SSIM of 0.9258, and an LPIPS of 0.0847 while maintaining an efficient inference time of 0.070 seconds. When refined with PnP flow matching, Moir\'eXNet achieves state-of-the-art results with a PSNR of 30.214 dB, surpassing DeMMamba~\cite{xu2024demmamba} (30.004 dB) by +0.21 dB. It also achieves the highest SSIM of 0.9281, outperforming VDRaw\cite{cheng2023recaptured} (0.9201) by +0.008, and achieves the lowest LPIPS score of 0.0795, which is 0.0054 lower than DeMMamba (0.0901).

These results demonstrate that the refined model leverages PnP flow matching to further enhance reconstruction quality. Both Moir\'eXNet variants excel in balancing computational efficiency and performance, with inference times competitive against lighter models. 
However, the increase in LPIPS alongside improved PSNR and SSIM after PnP flow matching refinement suggests that while the refinement enhances pixel-wise accuracy and structural fidelity, it may also introduce visually unnatural artifacts. Figure~\ref{fig:vdmoire} illustrates the effectiveness of our model in removing moiré patterns directly from RAW moiré images, without relying on the sRGB color space for color correction. The results demonstrate that our model not only eliminates moiré artifacts but also preserves accurate colors, avoiding any visible color distortions or artifacts.

For the TMM22 dataset, which focuses on image demoir\'eing, we compare Moir\'eXNet with state-of-the-art  methods, as presented in the table~\ref{table:quantitative-rawid}. As mentioned, the TTMM22 dataset only contains 540 images for training, which limits the model's ability to generalize effectively. Nevertheless, Moir\'eXNet achieves competitive performance against state-of-the-art methods, as illustrated in Figure~\ref{fig:TMM22}. 

\subsection{Ablation Study}
\noindent\textbf{1) Ablation study on the model architecture.}
Table~\ref{table:art} highlights the significance of each block in the MoiréXNet architecture. The full model, which integrates INN, LFEF, and TFMP, achieves the highest reconstruction quality, as reflected in the best PSNR and SSIM results. This underscores the necessity of combining these components to attain state-of-the-art performance on the VDraw dataset.

Specifically, including the INN block improves performance with a PSNR increase of +0.99 and an SSIM increase of +0.004. Adding the LFEF block on top of INN further enhances results, contributing a PSNR increase of +0.09 and an SSIM increase of +0.015. Finally, incorporating the TFMP block alongside INN and LFEF results in an additional PSNR increase of +0.09 and an SSIM increase of +0.003.

\begin{table}[htbp]
\caption{Ablation study on Model Architecture.}
\label{table:art}
\centering
    \scalebox{0.83}{    
        \begin{tabular}{lccc} 
            \toprule
            {Models}  & PSNR$\uparrow$ & SSIM$\uparrow$ \\\midrule
            Moir\'eXNet (w/o INN, LFEF and TFMP) & 29.04 & 0.906 \\
             Moir\'eXNet (w INN and w/o LFEF and TFMP) &  29.36 & 0.910  \\
            Moir\'eXNet (w INN and LFEF w/o TFMP) & 30.12 & 0.925 \\
            Moir\'eXNet (w INN LFEF and TFMP)  & \textbf{30.21} & \textbf{0.928} \\
            \bottomrule
        \end{tabular}}   
\end{table}

\begin{figure}[ht]
  \centering
  \includegraphics[width=0.8\linewidth]{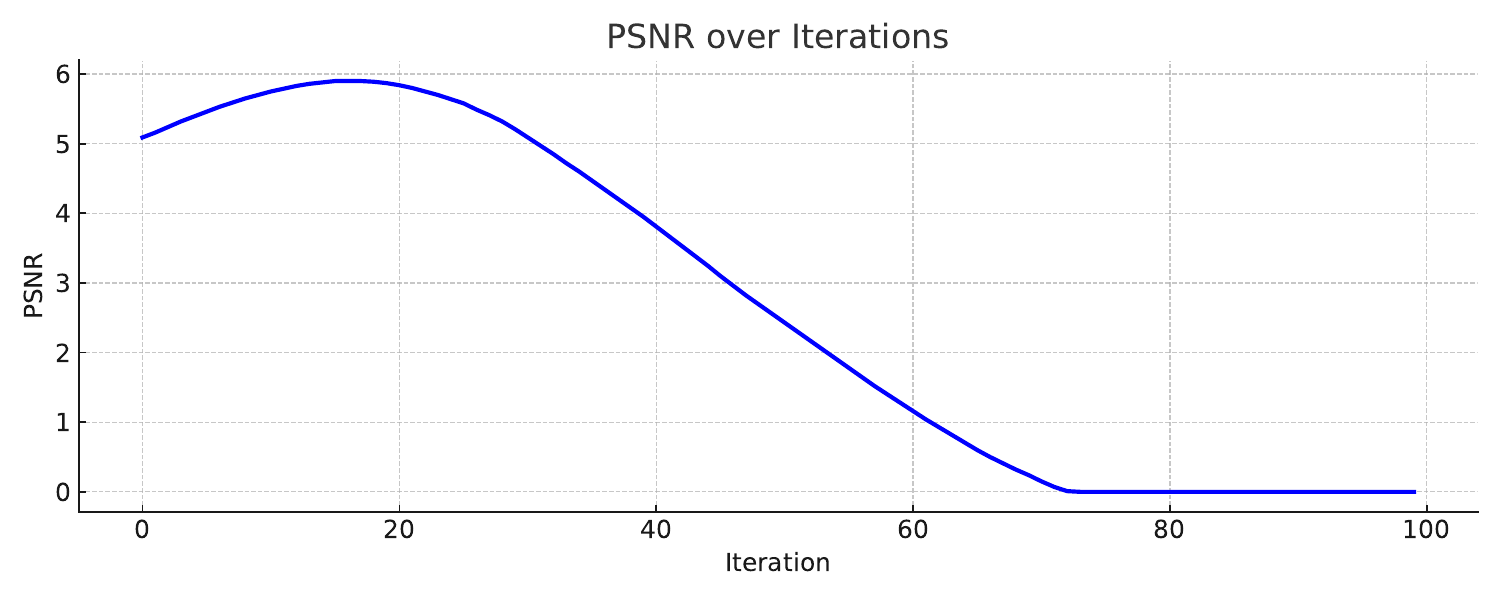}
  \caption{Denoiser iterations vs PSNR}
 \label{fig:denoiser_psnr_iteration}
\end{figure}

\noindent\textbf{2) Optimal t for Flow-Matching Denoising.} In flow-matching denoising, the parameter $t$ determines the progression of the algorithm toward the clean image. Since $\tilde{x}$ is already close to the clean image, fewer iterations are required. The Figure~\ref{fig:denoiser_psnr_iteration} demonstrates that the PSNR peaks around iteration 15, where the algorithm achieves optimal performance. This indicates that $\tilde{x}$ is approximately at $t=0.98$. To avoid overshooting the peak, we set $t=0.95$ for our method.

\section{Conclusion}
\label{sec:conc}

We proposed a hybrid approach for nonlinear moir\'{e} removal by combining an efficient supervised model with a denoising-based generative procedure, improving restoration quality and offering insights for handling linear and nonlinear degradations. Future work will focus on adaptive techniques to refine the data fidelity gradient, further enhancing the supervised model's performance.


\section*{Acknowledgments}
This should be a simple paragraph before the References to thank those individuals and institutions who have supported your work on this article.

\bibliographystyle{IEEEtran}
\bibliography{main}
 
%


 




\vfill

\end{document}